\documentclass{article}

\PassOptionsToPackage{numbers,sort&compress}{natbib}

\usepackage[numbers,sort&compress]{natbib}
\usepackage[final]{neurips_2025}
\usepackage{hyperref}

\usepackage[utf8]{inputenc}
\usepackage[T1]{fontenc}
\usepackage{xcolor}
\usepackage{formatting/packages}
\usepackage{formatting/hadi-colors}
\usepackage{formatting/hadi-macros-2025}
\usepackage{formatting/aliases}
\usepackage{formatting/formal-aliases}
\title{\compilert: LLM-Guided Optimizations for Efficient Model Serving}

\author{%
  Annabelle Sujun Tang \\
  University of California\\ San Diego \\
  \texttt{sujun@ucsd.edu}
  \And
  Christopher Priebe$^{*}$ \\
  University of California\\ San Diego \\
  \texttt{cpriebe@ucsd.edu}
  \And
  Rohan Mahapatra$^{*}$ \\
  University of California\\ San Diego \\
  \texttt{rohan@ucsd.edu}
  \AND
  Lianhui Qin \\
  University of California\\ San Diego \\
  \texttt{lianhui@ucsd.edu}
  \And
  Hadi Esmaeilzadeh \\
  University of California\\ San Diego \\
  \texttt{hadi@ucsd.edu}
}

\begin{document}

\maketitle

\begin{abstract}
While model serving has unlocked unprecedented capabilities, the high cost of serving large-scale models continues to be a significant barrier to widespread accessibility and rapid innovation. 
Compiler optimizations have long driven substantial performance improvements, but existing compilers struggle with neural workloads due to the exponentially large and highly interdependent space of possible transformations.
Although existing stochastic search techniques can be effective, they are often sample-inefficient and fail to leverage the structural context underlying compilation decisions.
We set out to investigate the research question of whether reasoning with large language models (LLMs), \hemph{without any retraining}, can leverage the context-aware decision space of compiler optimizations to significantly improve sample efficiency.
To that end, we introduce a novel compilation framework (dubbed \compiler) that formulates optimization as a sequential, context-aware decision process guided by a large language model and structured Monte Carlo tree search (MCTS).
The LLM acts as a proposal mechanism, suggesting hardware-informed transformations that reflect the current program state and accumulated performance feedback. 
MCTS incorporates the LLM-generated proposals to balance exploration and exploitation, facilitating a structured, context-sensitive traversal of the expansive compiler optimization space.
By achieving substantial speedups with markedly fewer samples than leading neural compilers, our approach demonstrates the potential of LLM-guided reasoning to transform the landscape of compiler optimization.\begingroup\hypersetup{hidelinks}\footnote{Code is available at {\url{https://github.com/he-actlab/REASONING_COMPILER}}}\endgroup
\def\thefootnote{*}\footnotetext{Equal contribution}
\end{abstract}

\section{Introduction}
\label{sec:introduction}
The rise of model serving for LLMs, diffusion models, and other neural models has enabled a new class of intelligent systems, driving transformative applications in healthcare, education, and scientific discovery.
These models incur significant computational demands during inference, which proportionally translate into substantial monetary costs.
Driving down the cost of model serving is critical, not merely to broaden access and democratize inference, but to catalyze faster cycles of innovation in model design and deployment.
Achieving this goal demands reducing inference runtime on computational infrastructure, resources that are not only expensive but also increasingly limited in availability.
Compiler optimizations are a critical enabler, not only for cost-efficient inferencing across diverse applications but also for empowering rapid research iteration.

Existing compilers struggle with neural models due to the exponentially large space of valid program transformations (e.g., tiling, fusion, and layout changes). 
Each decision, such as selecting a tiling factor or a parallelization strategy, introduces dependencies and constraints that influence the feasibility and performance benefits of subsequent transformations.
Rule-based optimizations also often rely on hand-tuned heuristics that can overfit to a specific workload or hardware target.
The seminal work in superoptimization~\cite{superoptimization:asplos:1987, denali:pldi:2002, peephole-superoptimizer:asplos:2006} aimed to tackle these shortcomings through enumerative or symbolic search, but the search space proved combinatorial and rugged.
STOKE~\cite{stoke:asplos:2013} showed that high-quality programs often lie in regions separated by low-probability paths, and therefore adopted Markov chain Monte Carlo (MCMC)-based randomized search.
Neural compilers followed suit, using evolutionary search or simulated annealing to navigate similarly irregular landscapes~\cite{tensor-comprehensions:arxiv:2018, autotvm:neurips:2020, ansor:osdi:2020, metaschedule:neurips:2022}.
While these methods have shown promise in discovering performant configurations, they are fundamentally sample-inefficient.
They overlook synergistic transformations that emerge only when decisions are made with contextual awareness.
These techniques also often explore redundant subspaces or invalid configurations.

In contrast, we set out to investigate the research question of whether reasoning with large language models (LLMs), \hemph{without any retraining}, can leverage the context-aware decision space of compiler optimizations to significantly improve sample efficiency.
To that end, we introduce a novel compilation framework that couples \hemph{LLM reasoning with Monte Carlo tree search (MCTS)~\cite{mcts}} to guide compiler optimization.
Hence, in our approach, compiler optimization is cast as \hemph{a sequential decision-making process}, in which each transformation, such as tiling, fusion, or vectorization, is selected with awareness of the current program state, while also \hemph{assimilating downstream information and propagating its implications upstream} to guide future decisions.
\hemph{Our approach avoids the prohibitive cost of fine-tuning LLMs as compilation policies, nor does it require additional training or task-specific adaptation.}
In this formulation, the LLM evaluates partial transformation sequences and proposes contextually appropriate next steps, drawing upon hardware-informed cost models and the historical trajectory of optimization decisions to inform its proposal.
The LLM serves as a context-aware proposal engine: given the current schedule and its observed performance, it generates candidate transformations that are likely to be effective in the context of the traversed trajectory.
These LLM-guided reasoning choices are integrated into an MCTS framework that provides a structured mechanism for balancing exploration and exploitation by evaluating LLM-suggested transformations, expanding promising branches, and leveraging rollout feedback to adaptively steer the search toward high-performing regions of the exponentially large optimization space.

This integration of LLM-based chain-of-thought (CoT) guidance with tree search \hemph{combines contextual reasoning and adaptability with principled, structured decision-making}, enabling the compiler to navigate the complexity of the search space with significantly improved sample efficiency.
We evaluate the \compiler and compare its improvements and sample efficiency with TVM, which employs evolutionary search.
Results show that the \compiler consistently achieves significantly higher speedups than what TVM achieves using significantly fewer samples.
On five representative benchmarks (\llamabench, \deepseekbench, \fluxattbench, \fluxconvbench, and \mlpbench) and across five hardware platforms (Amazon Graviton2, AMD EPYC 7R13, Apple M2 Pro, Intel Core i9, and Intel Xeon E3), the \compiler achieves 5.0$\times$ average speedup using 5.8$\times$ fewer samples, resulting in an average of 10.8$\times$ improvement over TVM in sample efficiency. 
For the end-to-end \llamathreeb benchmark across five hardware platforms, the \compiler uses 3.9$\times$ fewer samples to achieve a 4.0$\times$ speedup, yielding a 5.6$\times$ sample efficiency improvement.
These results underscore the promise of LLM-guided reasoning in neural compilation for efficient and scalable model serving.

\section{Problem Formalization}
\label{sec:formalization}
\begin{equation}
\mutatorseq_{\text{opt.}} = \argmax_{\strut\mathclap{\mutatorseq' \subseteq \mutatorset^*,\ |\mutatorseq'| \leq T}} \objfunc\!\left(\left(\mutator'_k \circ \cdots \circ \mutator'_1\right)\!\left(\program_0\right)\right)
\label{eq:problem}
\end{equation}
We consider the problem of optimizing an input program $\program_0 \in \programset$ representing a layer from a neural network for some objective function $\objfunc: \programset \mapsto \mathbb{R} \geq 0$.
This objective function represents an evaluation of the program on the target platform for some figure of merit (e.g., latency, power, utilization).
Any program $\program \in \programset$ can be transformed through the application of some transformation/optimization (used interchangeably from here on out) $\mutator \in \mutatorset$, where each optimization is a function $\mutator: \programset \mapsto \programset$ that performs a targeted transformation to the program, thus introducing a new variant of the program that is semantically equivalent to the original program but may perform better or worse on a target hardware platform.
In this way, successive application of transformations to a program can yield significant performance differences from the original.
Therefore, given some maximum transformation sequence length $T$, the goal is to find a sequence of transformations $\mutatorseq_{\text{opt.}} = \langle \mutator_1, \mutator_2, \ldots, \mutator_n \rangle$ such that $n \leq T$ and $\objfunc\!\left(\program_{\text{opt.}}\right) = \max_{\mutatorseq' \subseteq \mutatorset^*,\ |\mutatorseq'| \leq T} \objfunc\!\left(\left(\mutator'_k \circ \cdots \circ \mutator'_1\right)\!\left(\program_0\right)\right)$ where $\program_{\text{opt.}} = \left(\mutator_n \circ \mutator_{n-1} \circ \cdots \circ \mutator_1\right)\!\left(\program_0\right)$ and $\mutatorset^*$ is the Kleene star\footnote{The Kleene star operator, denoted with an asterisk ($^*$), represents the set of all finite-length sequences, including the empty sequence, formed from elements of a given set.} of $\mutatorset$.
These constraints collectively define the optimization objective given in \eref{eq:problem}.
To facilitate an efficient search over the space of valid program transformation sequences, we cast the optimization problem as a finite-horizon Markov decision process (MDP) defined by the tuple $\mathcal{M} = \langle \mathcal{S}, \mathcal{A}, \mathcal{P}, \mathcal{R} \rangle$.
This formulation provides a structured approach for sequential decision-making in the transformation space, allowing the search process to account for how individual transformations compound over time to affect final program performance.
Compared to unstructured methods such as exhaustive or purely stochastic search, which often require a large number of expensive program evaluations, casting the problem as an MDP enables more deliberate exploration, offering the potential for improved sample efficiency.
Each state $s_t \in \mathcal{S}$ corresponds to a program $\program_t \in \programset$ obtained by applying a sequence of transformations to the original program $\program_0$, i.e., $s_t = \program_t = \left(\mutator_t \circ \cdots \circ \mutator_1\right)\!\left(\program_0\right)$.
An action $a_t \in \mathcal{A}$ corresponds to selecting a transformation $\mutator \in \mutatorset$ to apply at step $t$, transitioning the current program to a new variant.
Since the application of a transformation is deterministic, the transition function $\mathcal{P}\!\left(s_{t+1} \mid s_t, a_t\right)$ is $1$ if $s_{t+1} = a_t\!\left(s_t\right)$ and 0 otherwise.
The reward function is defined as the objective value caused by the optimization sequence, i.e., $\mathcal{R}\!\left(s_t, a_t\right) = s \cdot  \objfunc\!\left(a_t\!\left(\program_t\right)\right)$ where $s \in \{+1,-1\}$ is chosen so that larger rewards are always better.
By formulating the problem as an MDP, we enable the use of planning algorithms such as Monte Carlo tree search (MCTS)~\cite{mcts} to explore program transformation sequences.
Under standard assumptions, such as finite branching, bounded rewards, and a tree policy (e.g., UCT) that guarantees persistent exploration, MCTS is consistent on finite-horizon problems: as the number of simulations tends to infinity, it converges (with probability 1) to the optimal root action/sequence $\mutatorseq_{\text{opt.}}$ that maximizes the objective.
With any finite simulation budget, it returns a high-quality but approximate solution. 
Consequently, our framework (see \sref{sec:framework}) yields a sequence $\mutatorseq_{\text{opt.}}’$ that approximately maximizes the objective in practice while enjoying asymptotic optimality in theory.

\section{\compilerheading: Integrating LLM-Guided Contextual Reasoning with Monte Carlo Tree Search}
\label{sec:framework}
We present the \compiler, a novel compilation framework that unifies the structured exploration capabilities of Monte Carlo tree search (MCTS)~\cite{mcts} with the contextual, history-aware reasoning of large language models (LLMs). 
While MCTS provides a principled approach to exploring sequences of program transformations, compiler optimization introduces a unique challenge: the successive application of transformations can exhibit complex, non-local interactions that are difficult to capture through purely stochastic or myopic policies.
To address this, we employ an LLM to model program transformation context, tracking which transformations have been applied, how they impact performance, and what directions remain promising. 
This contextualization is essential to enabling effective and sample-efficient search in compiler optimization.
\niparagraph{Optimization interactions are complex, making efficient search challenging.}
Unlike tasks where actions are relatively independent, program transformations compose in subtle and complex ways.
For example, the profitability of applying loop tiling may depend on the prior application of loop fusion or unrolling.
Additionally, transformations can introduce new, unforeseen opportunities/constraints for future transformations.
These dependencies make the space of valid and useful transformation sequences both combinatorial and deeply contextual.
While black-box methods such as evolutionary search and some implementations of reinforcement learning have achieved notable success in compiler autotuning~\cite{tvm, autotvm:neurips:2020, metaschedule:neurips:2022, chameleon:iclr:2020}, they often do not explicitly model the nuanced structural and temporal dependencies between transformations.
This can limit their ability to generalize across contexts, as optimization efficacy depends on transformation histories.
Even when guided by local reward signals, they may struggle to capture the interplay between past decisions and future opportunities, limiting their effectiveness in deeply contextual optimization landscapes.
Our insight is that efficient search in this space benefits from an agent that reasons over transformation history, structural code changes, and observed performance dynamics to choose the next step.

\begin{figure*}[!t]
    \centering
    \includegraphics[width=\linewidth]{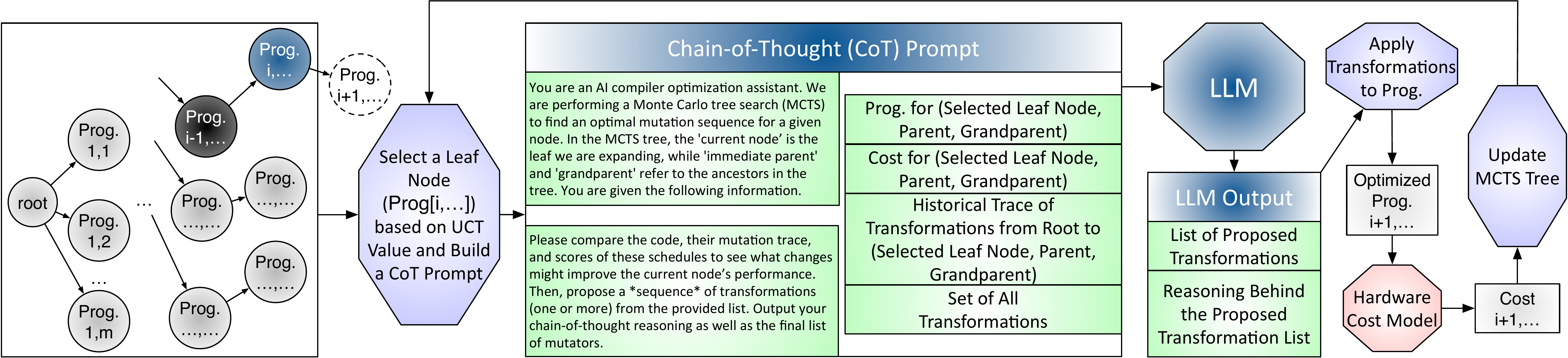}
    \caption{Overview of the optimization workflow. The algorithm explores the tree to select a candidate node. At this node, the LLM is prompted with contextual information to generate a sequence of transformations, which are then applied to produce optimized code variants.}
    \label{fig:framework}
\end{figure*}

\subsection{LLM-Guided Contextual Reasoning for Program Transformation Proposal}
\label{sec:llm-reasoning}
\niparagraph{Contextual reasoning via LLMs.} 
To address these challenges, the \compiler leverages a large language model (LLM) as a contextual reasoning engine. 
The LLM is tasked with synthesizing program transformations that are not only syntactically valid but also informed by the program's full history and structure. 
By prompting the LLM with a rich, structured representation of the current optimization state, we enable it to reason over the cumulative effects of prior transformations, analyze performance trends, and identify differential improvements over prior programs.
\fref{fig:framework} illustrates the optimization workflow. 
From the root program, the \compiler traverses the tree by computing the UCT score~\cite{uct:ecml:2006}, selecting a promising leaf node (i.e., program) $\program_i$ for expansion by balancing exploitation of high-reward paths and exploration of under-sampled branches based on visit statistics and empirical mean rewards (see \sref{sec:mcts}).
\niparagraph{Prompt construction.} 
At each expansion step in the search, the LLM receives a prompt that includes the source code and predicted performance for the current program $\program_i$, its parent $\program_{i - 1}$, and its grandparent $\program_{i - 2}$. 
It also includes the ordered sequences of transformations that were applied to reach each of these program variants, denoted $\mutatorseq_i$, $\mutatorseq_{i-1}$, and $\mutatorseq_{i-2}$. 
Finally, the full set of available transformation operations $\mutatorset$ is included.
Given this context, the LLM is explicitly instructed to: (1) analyze the differences between program variants and their associated performance scores, identifying which transformations contributed to observed performance changes; (2) reason about potential interactions between previously applied and candidate future transformations, including both synergistic and antagonistic effects; (3) synthesize a new sequence of transformations that is justified in the context of the current program structure and transformation history; and (4) provide a rationale for the proposed sequence, referencing specific code features and transformation interactions.
This structured prompt is designed to elicit chain-of-thought (CoT) reasoning~\cite{cot:neurips:2022}, encouraging the LLM to perform deep, multi-step analysis and move beyond surface-level edits, instead generating proposals that are both semantically meaningful and tailored to the evolving optimization trajectory.
\niparagraph{Transformation proposal and validation.} 
The LLM proposes a candidate transformation $\mutator_{i+1} \in \mutatorset$ in the form of a string.
Given the generative nature of the LLM, the output may include an invalid or unrecognized transformation even though it is guided by a predefined set of valid transformations.
To ensure correctness, the output string is first parsed and filtered to retain only a transformation that matches known valid names and transformation parameters.
If no valid transformation is found, the \compiler samples a random transformation from the valid set.
The successfully validated and applied transformation yields a new program variant $\program_{i+1}$, with its transformation history updated as $\mutatorseq_{i+1} = \mutatorseq_{i} \oplus \langle \mutator_{i+1} \rangle$, where $\oplus$ denotes sequence concatenation.
This new program variant is scored using a hardware cost model and used to update the MCTS tree (see \sref{sec:mcts}).
\hemph{It is important to emphasize that the LLM is not the centerpiece of our contribution, but a necessary enabler of effective search in this domain. 
Compiler optimization poses a uniquely challenging setting due to the non-local, compositional nature of transformation interactions. 
Traditional black-box search or heuristic-guided methods struggle to navigate such spaces efficiently. 
The \compiler uses structured search (via MCTS) with learned contextual reasoning (via LLM + CoT) to overcome these challenges. 
The result is a sample-efficient optimization algorithm capable of discovering performant transformation sequences in high-dimensional, high-interaction spaces.}
\begin{figure*}[!t]
    \centering
    \includegraphics[width=\linewidth]{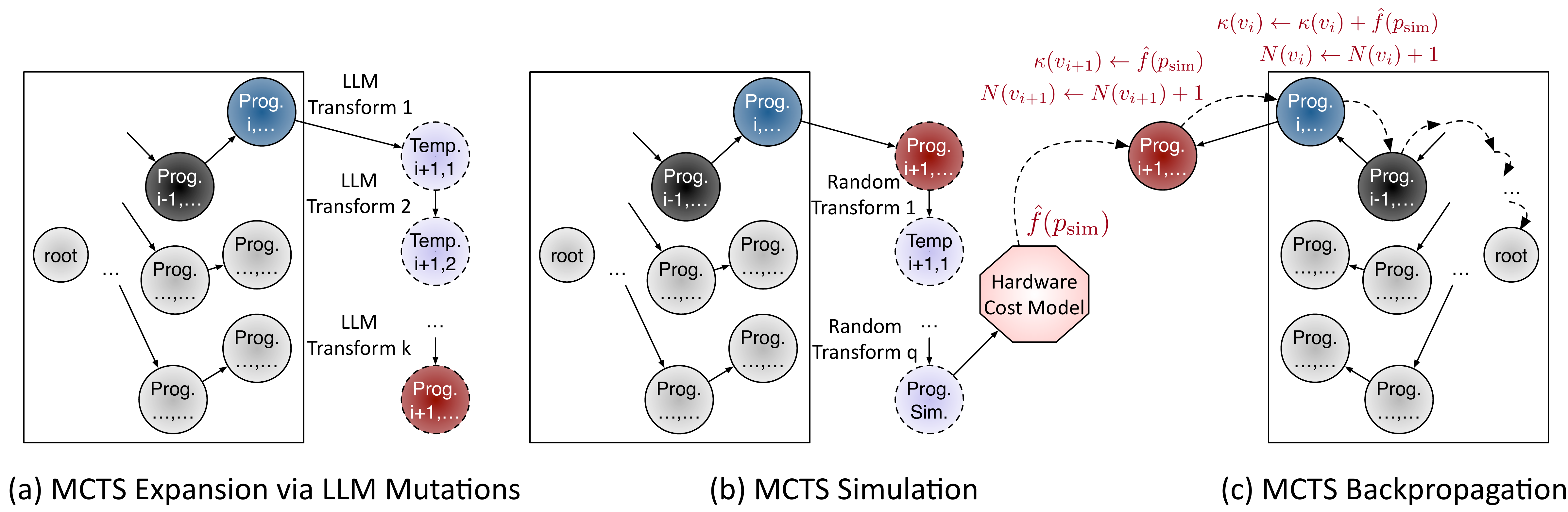}
    \caption{Structured tree search where nodes are (a) selected and expanded with the LLM suggested transformations, (b) scored by a learned hardware cost model, and (c) updated with performance estimates to guide future search.}
    \label{fig:tree-walk}
\end{figure*}
\subsection{Structured Optimization via Monte Carlo Tree Search}
\label{sec:mcts}
\niparagraph{MCTS as a sample-efficient planner.}
As described in \sref{sec:formalization}, we cast program optimization as a finite-horizon decision process over the space of transformation sequences. 
Framing the problem as an MDP allows the \compiler to consider long-term optimization effects and leverage planning algorithms such as Monte Carlo tree search (MCTS) to explore this space efficiently.
MCTS operates over a tree $\mathcal{T} = \langle V, E \rangle$ where $V = \programset$ and $E = \mutatorset$ such that each node $\program \in \programset$ is a program from the state space $\mathcal{S}$ and each edge $\mutator \in \mutatorset$ corresponds to a transformation from the action space $\mathcal{A}$. 
This tree structure naturally supports the reuse of common transformation prefixes and allows the planner to backpropagate value estimates from downstream program variants to upstream decisions.
Such reuse is critical in compiler optimization, where transformation sequences exhibit both compounding effects and long-range interactions.
\niparagraph{Selection via UCT.}
During the selection phase, MCTS traverses $\mathcal{T}$ from the root, recursively selecting child programs $\program_i$ to maximize the UCT (Upper Confidence bounds applied to Trees) criterion:
$$\text{UCT}\!\left(\program_i\right) = \frac{\mctscost\!\left(\program_i\right)}{N\!\left(\program_i\right)} + c \sqrt{\frac{\ln N\!\left(\program_{i-1}\right)}{N\!\left(\program_i\right)}}$$
where $\mctscost\!\left(\program_i\right)$ is the cumulative reward of $\program_i$, $N\!\left(\program_i\right)$ is the visit count of node (i.e., program) $\program_i$, and $c$ governs the exploration-exploitation tradeoff.
\niparagraph{LLM-guided expansion.}
As shown in \frefsub{fig:tree-walk}{a}, once a promising leaf node $\program_i$ is selected, an LLM is queried to propose a transformation conditioned on $\program_i$ and its ancestors (see \sref{sec:llm-reasoning}).
The model generates a candidate transformation $\mutator_{i+1} \in \mutatorset$, which is applied to $\program_i$ to produce a new program $\program_{i+1} = \mutator_{i+1}\!\left(\program_i\right)$.
This results in a new node $\program_{i+1}$ added to $\mathcal{T}$ corresponding to the updated program and extended transformation path.
To ensure $\mathcal{T}$ remains acyclic, if $\program_{i+1}$ already exists in the tree, it is not added.
By leveraging the LLM's contextual reasoning, the system proposes globally informed transformations that extend beyond myopic heuristics.
\niparagraph{Rollout for local reward estimation.}
As shown in \frefsub{fig:tree-walk}{b}, once a new node $\program_{i+1}$ is added to the tree, the \compiler performs a lightweight MCTS rollout to estimate the long-term impact of the transformation sequence that produced it.
This is done by sampling a randomized sequence of legal transformations $\mutator_1, \ldots, \mutator_q$ and applying them to obtain a terminal program $\program_{\text{sim}} = (\mutator_q \circ \cdots \circ \mutator_1)(\program_{i+1})$.
Directly measuring hardware-level performance requires compiling and running on real hardware, which is too expensive for the inner loop of a planning algorithm.
Following standard practice in compiler autotuning, the \compiler uses a learned, hardware-informed surrogate $\hat{f}$ for $\objfunc$ that is cheap to evaluate and accelerates search while preserving final quality~\cite{tvm,autotvm:neurips:2020,ansor:osdi:2020,dynatune:iclr:2021,glimpse:dac:2022,transfer-tuning:pact:2022}.
We convert this prediction into a rollout reward
$\mctscost\!\left(\program_{i+1}\right) = s \cdot \hat{f}\!\left(\program_{\text{sim}}\right)$, where
$s \in \{+1,-1\}$ is chosen so that larger values indicate better performance
(e.g., $s=-1$ for latency).
This noisy but informative proxy lets MCTS trade off immediate and downstream effects without real-hardware runs.

\niparagraph{Backpropagation.}
As shown in \frefsub{fig:tree-walk}{c}, the estimated reward $\mctscost\!\left(\program_{i+1}\right)$ is then backpropagated to all ancestors along the path to the root according to the update step $\mctscost\!\left(\program_A\right) \leftarrow \mctscost\!\left(\program_A\right) + \mctscost\!\left(\program_{i+1}\right)$ where $\program_A$ is some ancestor program.
The visit counts are also updated according to the update step $N\!\left(\program_A\right) \leftarrow N\!\left(\program_A\right) + 1$.
These updates refine the empirical estimates that guide future selections. 

\section{Evaluation}
\label{sec:results}
We implement the \compiler as an extension to MetaSchedule~\cite{metaschedule:neurips:2022}.
The framework introduces three modular components: (1) a prompt generator that serializes the current scheduling state, including the IRModule, transformation trace (i.e., the applied schedule history), and hardware cost model outputs, into structured prompts that capture the textual difference from the base IRModule and reflect the current schedule’s performance; (2) an LLM interface that queries an external API (e.g., OpenAI) and parses the LLM's output into candidate transformation sequences; and (3) a tree manager that performs MCTS with selection based on UCT score, expansion using LLM suggested transformations, simulation with a hardware-informed cost model, and backpropagation for tree statistics updates.
\subsection{Methodology}
\label{sec:experimental_setup}
We evaluate the \compiler on five representative computational kernels drawn from production-scale models: (1) a self-attention layer from \llamathreeb~\cite{llama:arxiv:2024}, (2) a mixture-of-experts (MoE) layer from \deepseekrone~\cite{deepseek:arxiv:2025}, (3) a self-attention layer from \flux (stable diffusion)~\cite{flux:gitrepo:2024}, (4) a convolution layer from \flux~\cite{flux:gitrepo:2024}, and (5) an MLP layer from \llamafourscout~\cite{llama4:huggingface:2025}.
In addition, we perform an end‑to‑end evaluation of \llamathreeb.
Compiler optimization is framed as a sequential decision process and guided by MCTS~\cite{mcts} using the Upper Confidence bounds applied to Trees (UCT) criterion~\cite{uct:ecml:2006} with exploration parameter $c = \sqrt{2}$ and branching factor $B = 2$ following prior work~\cite{branching1:cg:2007, ucb:ml:2002}. 
During search, the LLM (OpenAI GPT-4o mini~\cite{openai:4o-mini:2025}) is queried using hierarchical context---specifically, the parent and grandparent schedules and their transformations---to enable informed proposal generation. 
We compare three optimization strategies: (1) TVM  MetaSchedule~\cite{metaschedule:neurips:2022}, which uses \baseline; (2) MCTS without LLM guidance (\mcts); and (3) the \compiler that uses prompt-based proposal generation (\llmmcts). 
All experiments are conducted using Apache TVM v0.20.0~\cite{tvm, tvm:v0200:2025}. Our experimental environment is a dedicated Intel Core i9 workstation under a fixed software and hardware stack to isolate scheduling effects. This environment covers all five kernels above and is the ablation environment. To show portability and scalability across consumer and datacenter processors, we evaluate each of the five kernels on five hardware platforms: Amazon Graviton2, AMD EPYC 7R13, Apple M2 Pro, Intel Core i9, and Intel Xeon E3.
Each experiment is repeated 20 times, and we report the mean performance to ensure statistical stability.
Additionally, we leverage OpenAI and HuggingFace model serving APIs to access the respective models.
The implementation is open-sourced.

\begin{figure*}[!h]
    \centering
    \includegraphics[width=\linewidth]{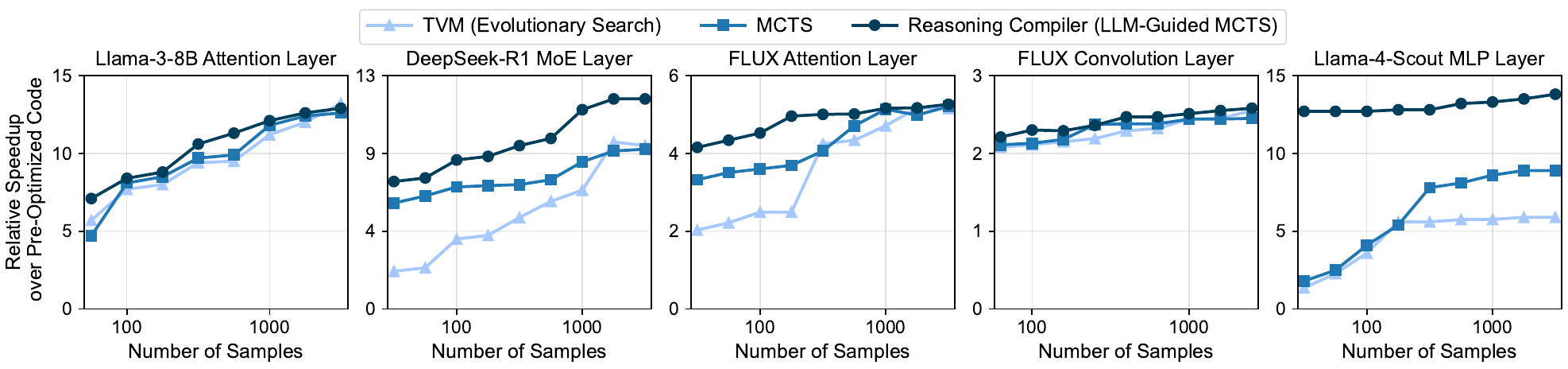}
    \caption{Relative speedup over pre-optimized code as a function of evaluated transformation proposals. The \compiler achieves superior sample efficiency, discovering high-quality code with fewer samples across all operators in low-budget regimes.}
    \label{fig:improvement_a}
\end{figure*}

\subsection{Experimental Results}
\label{sec:results:speedup}

We assess the sample efficiency of the \compiler by analyzing how code quality evolves with increasing search budget, quantified in terms of evaluated transformation proposals. 
\fref{fig:improvement_a} presents results across five representative workloads, encompassing both transformer-style attention layers and convolution-heavy architectures. 
Across all benchmarks, our method achieves competitive or superior code performance with significantly fewer samples than state-of-the-art black-box autotuners such as TVM with \baseline. 
These results directly support the central hypothesis of our work: \hemph{leveraging LLM-driven, context-aware reasoning enables more efficient and effective exploration of the compiler optimization space}.

\niparagraph{Rapid convergence in low-sample regimes.} 
A consistent trend across all benchmarks is the rapid ascent of code quality in the initial stages of search. 
This early-stage performance is critical in practice, as real-world compiler pipelines often operate under strict tuning time budgets. 
\fref{fig:improvement_a} shows \emph{Relative Speedup over Pre-Optimized Code} on the y-axis, with the number of evaluated transformation proposals on the x-axis. 
Speedup is defined as the ratio of the execution time of the unoptimized code to that of the optimized code after tuning. 
Higher values indicate more efficient and optimized code. 
For instance, on the \llamathreeb self-attention layer, the \compiler achieves a 7.08$\times$ speedup over the untuned baseline with just 36 samples, whereas TVM with \baseline requires 72 samples, which is twice the budget to achieve comparable gains. 
On the \llamafourscout MLP layer, the gap is even more pronounced: the \compiler achieves 12.7$\times$ speedup at 20 samples, while TVM falls short of this mark even after 3000 samples. 

\niparagraph{Quantitative sample efficiency.} 
To formally quantify sample efficiency, we compare the number of samples required by each method to reach target speedups. 
On the \flux self-attention layer, the \compiler attains a 2$\times$ speedup using only 36 samples, while TVM with \baseline requires more than 600 samples, a 16$\times$ reduction in tuning cost. 
On the \flux convolution layer, the \compiler consistently outperforms TVM across nearly all budget levels and reaches TVM’s final performance after evaluating just 400 samples. 

\begin{table}[!b]
  \centering
  \small
  \renewcommand{\arraystretch}{0.9}
  \setcellgapes{-2.5pt}
  \makegapedcells
  \setlength{\tabcolsep}{1.6pt} 
  \caption{Sample efficiency comparison between the \compiler and TVM with \baseline on layer-wise benchmarks across various hardware platforms.}
  \label{tab:performance-layerwise}
  \begin{tabular}{@{}llrrrrrr@{}} 
  \toprule
  \multicolumn{1}{c}{\multirow{2}{*}[-2.5ex]{\textbf{\makecell{Hardware \\ Platform}}}} &
\multicolumn{1}{c}{\multirow{2}{*}[-2.6ex]{\textbf{Benchmark}}} &
\multicolumn{2}{c}{\textbf{TVM}} &
\multicolumn{2}{c}{\compilertable} &
\multicolumn{2}{c}{\textbf{Improvement}} \\
\cmidrule(lr){3-4} \cmidrule(lr){5-6} \cmidrule(lr){7-8}
& & \textbf{\# Samples} & \textbf{Speedup} &
      \textbf{\# Samples} & \textbf{Speedup} &
      \textbf{\makecell{Sample \\ Reduction}} &
      \textbf{\makecell{Sample \\ Efficiency \\ Gain}} \\
    \midrule
    \multirow{5}{*}{\textbf{\makecell[c]{Amazon \\ Graviton2}}} 
                          & \llamathreebtable Layer & 510  & 3.9$\times$ & 60   & 5.1$\times$ & 8.5$\times$  & 11.1$\times$ \\
                          & \deepseekronetable MoE Layer    & 980  & 2.7$\times$ & 150  & 5.9$\times$ & 6.5$\times$  & 14.4$\times$ \\
                          & \fluxtable Attention Layer     & 320  & 1.6$\times$ & 130  & 3.0$\times$   & 2.5$\times$  & 4.6$\times$  \\
                          & \fluxtable Convolution Layer   & 160  & 1.8$\times$ & 20   & 4.1$\times$ & 8.0$\times$  & 18.2$\times$ \\
                          & \llamafourscouttable MLP Layer  & 1,630 & 1.7$\times$ & 500  & 4.0$\times$ & 3.3$\times$  & 7.7$\times$  \\
    \midrule
    \multirow{5}{*}{\textbf{\makecell[c]{AMD \\ EPYC 7R13}}} 
                          & \llamathreebtable Layer & 1,400 & 2.1$\times$ & 200  & 12.1$\times$ & 7.0$\times$  & 40.3$\times$ \\
                          & \deepseekronetable MoE Layer    & 2,290 & 1.7$\times$ & 330  & 2.3$\times$  & 6.9$\times$  & 9.4$\times$  \\
                          & \fluxtable Attention Layer     & 2,460 & 1.5$\times$ & 230  & 3.1$\times$  & 10.7$\times$ & 22.1$\times$ \\
                          & \fluxtable Convolution Layer   & 2,520 & 1.3$\times$ & 470  & 4.8$\times$  & 5.4$\times$  & 19.6$\times$ \\
                          & \llamafourscouttable MLP Layer  & 510  & 6.4$\times$ & 100  & 10.2$\times$ & 5.1$\times$  & 8.1$\times$  \\
    \midrule
    \multirow{5}{*}{\textbf{\makecell[c]{Apple \\ M2 Pro}}} 
                          & \llamathreebtable Attention Layer & 1,010 & 3.3$\times$ & 190  & 9.7$\times$  & 5.3$\times$  & 15.6$\times$ \\
                          & \deepseekronetable MoE Layer    & 1,040 & 2.8$\times$ & 230  & 4.8$\times$  & 4.5$\times$  & 7.8$\times$  \\
                          & \fluxtable Attention Layer     & 270  & 2.1$\times$ & 50   & 3.7$\times$  & 5.4$\times$  & 9.5$\times$  \\
                          & \fluxtable Convolution Layer   & 2,260 & 1.5$\times$ & 510  & 5.5$\times$  & 4.4$\times$  & 16.2$\times$ \\
                          & \llamafourscouttable MLP Layer  & 2,460 & 2.2$\times$ & 440  & 3.4$\times$  & 5.6$\times$  & 8.6$\times$  \\
    \midrule
    \multirow{5}{*}{\textbf{\makecell[c]{Intel \\ Core i9}}} 
                          & \llamathreebtable Attention Layer & 920  & 10.5$\times$ & 130  & 11.0$\times$   & 7.1$\times$  & 7.4$\times$  \\
                          & \deepseekronetable MoE Layer    & 1,632 & 9.1$\times$  & 192  & 9.1$\times$  & 8.5$\times$  & 8.5$\times$  \\
                          & \fluxtable Attention Layer     & 1,000 & 5.1$\times$  & 150  & 5.4$\times$  & 6.7$\times$  & 7.0$\times$  \\
                          & \fluxtable Convolution Layer   & 400  & 2.3$\times$  & 72   & 2.3$\times$  & 5.6$\times$  & 5.6$\times$  \\
                          & \llamafourscouttable MLP Layer  & 230  & 5.6$\times$  & 20   & 12.7$\times$ & 11.5$\times$ & 26.1$\times$ \\
    \midrule
    \multirow{5}{*}{\textbf{\makecell[c]{Intel \\ Xeon E3}}} 
                          & \llamathreebtable Attention Layer & 2,760 & 3.9$\times$ & 320  & 5.8$\times$  & 8.6$\times$  & 12.8$\times$ \\
                          & \deepseekronetable MoE Layer    & 1,000 & 3.7$\times$ & 180  & 4.4$\times$  & 5.6$\times$  & 6.6$\times$  \\
                          & \fluxtable Attention Layer     & 1,340 & 1.4$\times$ & 450  & 3.4$\times$  & 3.0$\times$  & 7.1$\times$  \\
                          & \fluxtable Convolution Layer   & 220  & 1.9$\times$ & 40   & 2.2$\times$  & 5.5$\times$  & 6.4$\times$  \\
                          & \llamafourscouttable MLP Layer  & 1,200 & 2.0$\times$ & 300  & 6.1$\times$  & 4.0$\times$  & 12.2$\times$ \\
    \midrule
    \makecell[c]{\textbf{Geomean}}      & ---  & ---   & \textbf{2.7$\times$} & ---  & \textbf{5.0$\times$} & \textbf{5.8$\times$}  & \textbf{10.8$\times$} \\
    \bottomrule
  \end{tabular}
\end{table}

\niparagraph{Speedup relative to baselines.} 
The \compiler not only produces better code, but does so \emph{more aggressively and earlier} in the search process. 
For example, on the \deepseekbench, the \compiler achieves a 3.3$\times$ speedup over TVM with \baseline at 36 samples; on the \llamafourscout MLP layer, the \compiler achieves a 9.3$\times$ speedup over TVM at 20 samples.
This trend, which shows strong initial gains followed by convergence, demonstrates that the \compiler quickly identifies high-performing regions of the search space, while TVM’s uninformed search requires substantial exploration to reach similar quality. 

\niparagraph{Operator-specific trends.} 
We observe that certain operator types, such as matrix multiplication operations extracted from attention layers and MLP layers, exhibit sharper performance improvements. 
This is likely due to recurring structural patterns such as loop fusion, tiling, and vectorization, which pretrained LLMs can more readily recognize and exploit. 
Convolutional operators, by contrast, expose a broader and less regular transformation space. 
Nonetheless, the \compiler consistently matches or exceeds baseline performance with fewer samples, underscoring its effectiveness across diverse operator characteristics. 

\niparagraph{Sample efficiency across hardware platforms.} 
As shown in \tref{tab:performance-layerwise}, the \compiler demonstrates superior sample efficiency compared to TVM with \baseline across five hardware platforms on five benchmarks. 
We define sample efficiency as the speedup achieved per sample ($\frac{Speedup}{\#~of~Samples}$). 
On average, across all 25 platform-operator pairs, the \compiler achieves a 5.0$\times$ speedup using 5.8$\times$ fewer samples, resulting in a 10.8$\times$ improvement in sample efficiency. 
The performance gains are particularly significant for compute-intensive workloads. 
For instance, for the \llamathreeb self-attention layer on AMD EPYC 7R13, the \compiler achieved a 12.1$\times$ speedup in just 200 samples, while TVM required 1,400 samples to reach a 2.1$\times$ speedup. 
This represents a 7.0$\times$ sample reduction and a 40.3$\times$ sample efficiency gain. 
On Intel Core i9, the \compiler often matches or exceeds TVM's peak with fewer trials: on the \llamafourscout MLP layer, the \compiler used 11.5$\times$ fewer samples for a 26.1$\times$ efficiency gain.

\begin{table}[ht]
  \centering
  \small 
  \setcellgapes{-2.5pt}
  \renewcommand{\arraystretch}{0.9}
  \setlength{\tabcolsep}{8.5pt} 
  \caption{Sample efficiency comparison between the \compiler and TVM with \baseline on end-to-end \llamathreeb across various hardware platforms.}
  \label{tab:performance-comparison-final}
  \begin{tabular}{@{}lrrrrrr@{}}
    \toprule
    
    \multicolumn{1}{c}{\multirow{2}{*}[-3.5ex]{\textbf{\makecell{Hardware \\ Platform}}}} & 
    \multicolumn{2}{c}{\textbf{TVM}} & 
    \multicolumn{2}{c}{\compilertable} & 
    \multicolumn{2}{c}{\textbf{Improvement}} \\
    
    \cmidrule(lr){2-3} \cmidrule(lr){4-5} \cmidrule(lr){6-7}
    
    & 
    \textbf{\# Samples} & \textbf{Speedup} & 
    \textbf{\# Samples} & \textbf{Speedup} & 
    \textbf{\makecell{Sample \\ Reduction}} & 
    \textbf{\makecell{Sample \\ Efficiency \\ Gain}} \\
    
    \midrule
    \makecell[c]{\textbf{Amazon Graviton2}}    & 4,560 & 3.7$\times$ & 1,440 & 5.1$\times$ & 3.2$\times$ & 4.4$\times$ \\
    \makecell[c]{\textbf{AMD EPYC 7R13}}       & 410   & 2.0$\times$ & 140   & 2.2$\times$ & 2.9$\times$ & 3.2$\times$ \\
    \makecell[c]{\textbf{Apple M2 Pro}}        & 4,820 & 2.2$\times$ & 1,770 & 3.9$\times$ & 2.7$\times$ & 4.8$\times$ \\
    \makecell[c]{\textbf{Intel Core i9}}       & 3,800 & 2.2$\times$ & 720   & 4.9$\times$ & 5.3$\times$ & 11.8$\times$ \\
    \makecell[c]{\textbf{Intel Xeon E3}}       & 4,640 & 5.0$\times$ & 670   & 5.0$\times$ & 6.9$\times$ & 6.9$\times$ \\
    \midrule
    \makecell[c]{\textbf{Geomean}}      & ---   & \textbf{2.8$\times$} & ---   & \textbf{4.0$\times$} & \textbf{3.9$\times$} & \textbf{5.6$\times$} \\
    \bottomrule
  \end{tabular}
\end{table}

\niparagraph{End-to-end sample efficiency.} For end-to-end \llamathreeb across the five hardware platforms in \tref{tab:performance-comparison-final}, the \compiler's sample efficiency improvement over TVM ranges from 3.2$\times$ on AMD EPYC to 11.8$\times$ on Intel Core i9. 
End-to-end speedups range from 2.2$\times$ on AMD EPYC to 5.1$\times$ on Amazon Graviton2. 
The \compiler consistently achieves significantly higher speedups: using 3.9$\times$ fewer samples, it achieves a 4.0$\times$ speedup and yields a 5.6$\times$ geometric-mean sample efficiency improvement over TVM with \baseline.

\niparagraph{Implications.} 
These findings reinforce our core thesis: compiler optimization should be cast as a structured decision process, enriched by prior knowledge and contextual reasoning. 
Our integration of LLMs into Monte Carlo tree search results in a strategically guided and sample-efficient search, particularly valuable in scenarios with constrained tuning budgets. 
By generating performant code with orders-of-magnitude fewer samples, our framework offers both practical deployment advantages and a compelling alternative to conventional, sample-inefficient compilation pipelines. 

\begin{figure*}[!h]
    \centering
    \includegraphics[width=\linewidth]{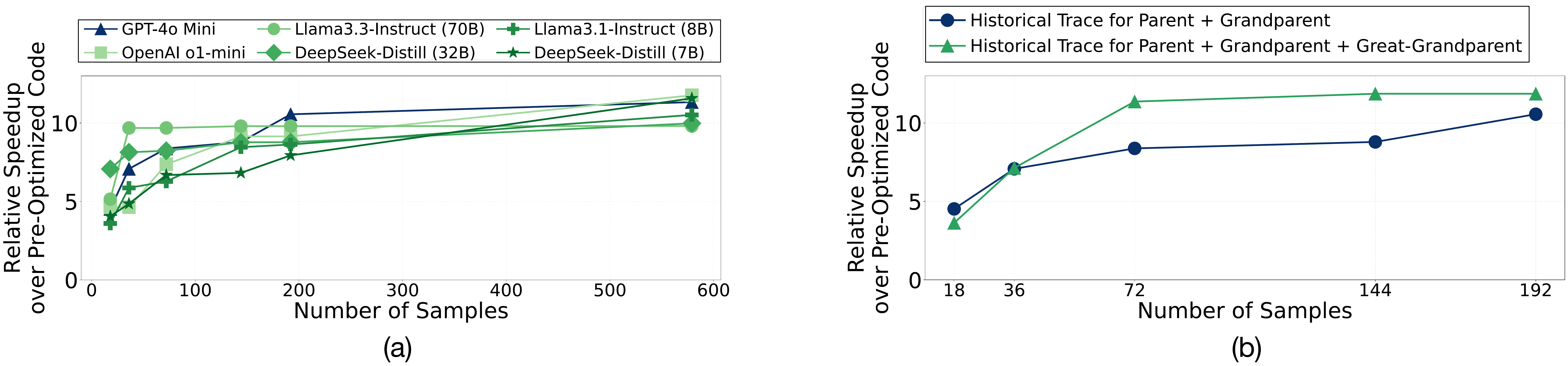}
    \caption{Ablation studies for the \llamathreeb self-attention layer. (a) Comparing different LLMs as proposal engines shows stronger LLMs lead to faster convergence. (b) Increasing the prompt's historical trace depth improves sample efficiency.}
    \label{fig:llm-ablation}
\end{figure*}

\subsection{Ablation Studies}
\subsubsection{Impact of LLM Choice and Reasoning Strategy}
\label{sec:llm-ablation}
To better understand the contributions of different components in our approach, we conduct an ablation study focused on the effects of LLM selection and reasoning modality.
\frefsub{fig:llm-ablation}{a} shows the relative speedup over unoptimized code as a function of the number of schedule samples evaluated by the \compiler on the \llamathreeb self-attention layer using a range of LLM models for API calls.
The x-axis indicates the cumulative number of schedules explored, while the y-axis shows the best speedup achieved so far. This setup enables us to directly compare how effectively various LLMs leverage contextual information to guide the search.
The general trend of the results supports our central claim: compiler optimization benefits from goal-directed, context-aware reasoning in terms of sample efficiency.
Below, we discuss the specific behaviors that exemplify different reasoning strategies.

\niparagraph{Large instruction-tuned Llama3.3 (70B) achieves exceptional sample efficiency}. The instruction-tuned Llama3.3-70B model rapidly attains near-optimal performance, reaching a 9.69$\times$ speedup after only 36 samples, roughly 86\% of the GPT-4o mini’s maximum speedup but with less than 6\% of its sampling budget. 
This corresponds to an approximately 15$\times$ improvement in sample efficiency.
Instruction tuning also significantly improves the ability of LLMs to generate domain-specific, context-aware transformation proposals. 
The consistent performance advantage of instruction-tuned models over untuned counterparts of comparable size confirms that semantic task alignment, combined with sufficient model capacity, synergistically enhances the effectiveness of sequential context reasoning in guiding compiler optimizations.

\niparagraph{DeepSeek-R1-Distill-Qwen (32B) excels in long-horizon optimization.}
The DeepSeek-R1-Distill-Qwen-32B model, employing a mixture-of-experts (MoE) architecture, exhibits a more gradual improvement, starting with a 7.07$\times$ speedup at 18 samples and reaching 9.98$\times$ after 579 samples. 
The sparse expert routing inherent in MoE architectures likely facilitates exploration of complex transformation sequences over extended horizons, complementing context-aware reasoning by enabling specialized and conditional decision-making.
\niparagraph{Lower-parameter models also achieve high sample efficiency.}
Despite their reduced scale, smaller models still produce notable speedups relative to the untuned baseline. 
For example, Llama3.1-Instruct (8B) reaches a 5.87$\times$ speedup, and DeepSeek-R1-Distill-Qwen (7B) achieves a 4.86$\times$ speedup at just 36 samples. 
When compared to the widely used \baseline strategy, which requires around 72 samples to achieve a 7.0$\times$ speedup and fails to reach comparable performance for tuning the \deepseekrone MoE layer.
Even after 3000 samples, these smaller models consistently outperform.
\hemph{The \compiler with lower-parameter models achieves at least twice the sample efficiency of TVM with \baseline, making them well-suited for efficient compiler optimization in local or edge deployments.}
\niparagraph{Open-source models match proprietary models in performance.}
Our results demonstrate that open-source LLMs, when adequately scaled and instruction-tuned, match or exceed the performance of proprietary baselines such as GPT-4o mini.
This underscores the broad applicability of our approach and its independence from proprietary data or architectures, enabling widespread adoption of context-aware, LLM-guided compiler optimization.

\subsubsection{Impact of Historical Trace Depth on Optimization Efficiency}
\frefsub{fig:llm-ablation}{b} presents the relative speedup over unoptimized code as a function of the number of schedule samples evaluated by the \compiler on the \llamathreeb self-attention layer. 
Using a deeper historical trace (see~\fref{fig:framework}) in the prompt (parent + grandparent + great-grandparent) leads to faster convergence compared to the shallower trace (parent + grandparent).
For example, at 36 samples, the deeper trace achieves a speedup of approximately 7.13$\times$, slightly surpassing the 7.08$\times$ of the shallower trace. 
However, by 72 samples, the deeper trace saturates at 11.36$\times$ speedup, while the shallower trace reaches only 8.38$\times$, requiring many more samples (around 579) to approach  11.3$\times$ performance.
\hemph{This demonstrates that including longer historical context enables the LLM to better capture dependencies and synergies in transformation sequences, resulting in more sample-efficient and goal-directed exploration, validating the advantage of context-aware reasoning.}

\section{Related Work}
\label{sec:related_work}
\niparagraph{Superoptimization.}
While our high-level goal of discovering highly efficient program variants shares motivation with the
superoptimization literature, our formulation and tractability differ substantially.
Superoptimization aims to find the globally optimal instruction-level program, typically via enumerative~\cite{superoptimization:asplos:1987, peephole-superoptimizer:asplos:2006}, symbolic~\cite{denali:pldi:2002}, or stochastic~\cite{stoke:asplos:2013} search over low-level assembly variants; hybrid~\cite{lens:asplos:2016} and neural~\cite{learning-superoptimization:iclr:2017} approaches have also been explored.
STOKE~\cite{stoke:asplos:2013} showed that high-quality programs often reside in low-probability regions and made the leap to use randomized search (MCMC).
Neural compilers followed suit and relied on evolutionary search or simulated annealing algorithms~\cite{tensor-comprehensions:arxiv:2018, autotvm:neurips:2020, ansor:osdi:2020, metaschedule:neurips:2022}.
In contrast, the \compiler treats optimization as a planning problem that leverages MCTS to reason contextually about dependencies among transformations over structured intermediate representations.

\niparagraph{ML-Based Autotuning.} 
Autotuning frameworks optimize performance-critical parameters (e.g., loop tile sizes, phase orderings, memory layouts) using a variety of ML-based techniques, including linear models~\cite{pred-unroll-factor:cgo:2005, micomp:taco:2017}, tree-based methods~\cite{auto-construct-inlining-heuristic:cgo:2013, protuner:arxiv:2020}, Bayesian networks~\cite{cobayn:taco:2016, baco:asplos:2023}, evolutionary algorithms~\cite{auto-construct-inlining-heuristic:cgo:2013, mlgo:arxiv:2021, kmeans-auto-tuning:cgo:2025}, clustering~\cite{micomp:taco:2017, kmeans-auto-tuning:cgo:2025}, and reinforcement learning~\cite{chameleon:iclr:2020, autophase:mlsys:2020, mlgo:arxiv:2021, kmeans-auto-tuning:cgo:2025}.
The \compiler shares the same goal of performance-driven parameter selection, but distinguishes itself by combining LLM-based contextual reasoning with structured search (via MCTS) to explore transformation sequences in a history- and structure-aware manner.
\niparagraph{Techniques for Neural Compilation.}
A large body of work targets the optimization of neural network inference pipelines, spanning graph-level transformations and scheduling~\cite{dl-with-dynamic-graph:arxiv:2017, taso:sosp:2019, neocpu:atc:2019, go:neurips:2020, ios:mlsys:2021, fusionstitching:arxiv:2021, equality-saturation-tensor-graph-superopt:mlsys:2021, apollo:mlsys:2022} and low-level code generation~\cite{tensor-comprehensions:arxiv:2018, tvm, tiramisu:cgo:2019, fireiron:pact:2020, unit:cgo:2021, dse-cnn:asplos:2021, deepcuts:pldi:2021, hidet:asplos:2023}.
Many modern systems incorporate learned components (e.g., cost models) and search strategies to navigate large configuration spaces; for example, TVM/Ansor~\cite{tvm, autotvm:neurips:2020, ansor:osdi:2020} and FlexTensor~\cite{flextensor:asplos:2020} use learned performance models and evolutionary strategies for tuning.
While highly effective for tensor-program tuning, these approaches often emphasize local parameter optimization or rely on domain-specific heuristics.
The \compiler moves beyond these works by introducing LLM-based contextual reasoning over transformation history, structural changes, and performance trends, enabling history- and structure-aware exploration not addressed in prior neural compilation work.
\niparagraph{LLMs for Code Reasoning and Optimization.}
LLMs have demonstrated capabilities in code generation~\cite{codellama, starcoder, magicoder, codegeex, codet5, deepseekcoder}, fuzzing~\cite{llm-fuzzing}, bug repair~\cite{prog-repair-llm}, and even high-level optimization~\cite{pie:iclr:2024}. 
Recent work has explored the use of LLMs to generate phase orderings or perform disassembly~\cite{llm-compiler-optimization:arxiv:2023, llm-compiler:cc:2025}.
The \compiler advances these approaches by embedding an LLM in a structured decision loop, leveraging it for context-aware reasoning within a grounded search process.
\section{Conclusion}
\label{sec:conclusion}
Compiling neural workloads remains a bottleneck for scalable model serving: traditional compilers struggle with combinatorial transformation spaces, and the state-of-the-art neural compilers rely on stochastic search, lacking sample efficiency and contextual awareness.
This paper introduced the \compiler, a novel framework that formulates compiler optimization as a sequential, context-aware decision process, pairing LLM‑generated proposals with MCTS and performance feedback to reason and navigate through the optimization space efficiently. 
By enabling LLM reasoning in the compiler optimization process, we achieve a leap from randomized search to informed and guided compilation.
Our results show that the \compiler consistently yields faster runtimes with markedly fewer evaluations without any retraining.
These gains directly translate to reduced operational cost of LLM services, lower energy usage per query, improved system responsiveness, more agile model deployment, faster model training, and accelerated innovation cycles, among other benefits.
Looking ahead, the same LLM that guides compilation can accelerate its own inferencing, creating a virtuous, self‑optimizing cycle in which sped-up LLMs enable more efficient transformations and progressively better models and services.

\section*{Acknowledgments}
We thank the anonymous reviewers for their insightful feedback. This work was in part supported by the National Science Foundation (NSF) award CCF \#2107598. The U.S. Government is authorized to reproduce and distribute reprints for governmental purposes not withstanding any copyright notation thereon. The views contained herein are those of the authors and should not be interpreted as necessarily representing the official policies or endorsements, either expressed or implied by the U.S. Government.
\bibliographystyle{unsrtnat}
\bibliography{bib/full_ref}

\newpage
\appendix
\section{LLM Prompt Example}
Below we show an example prompt used in our \llmmcts framework (refer to Figure~\ref{fig:framework}).

\textbf{Example Code to be Optimized:}
\begin{verbatim}
@tvm.script.ir_module
class MyModule:
   @T.prim_func
   def main(                                    
      A: T.Buffer((1, 16, 7168), "float32"),
      B: T.Buffer((7168, 2048), "float32"),
      C: T.Buffer((1, 16, 2048), "float32"),
      ):
      T.func_attr({"global_symbol": "main", "tir.noalias": True})
      for b, t, j, k in T.grid(1, 16, 2048, 7168):
         with T.block("moe"):
            vb, vt, vj, vk = T.axis.remap("SSSR", [b, t, j, k])
            with T.init():
               C[vb, vt, vj] = T.float32(0)
            C[vb, vt, vj] += A[vb, vt, vk] * B[vk, vj]
\end{verbatim}

\begin{promptbox}[Prompt]
You are a code optimization assistant performing Monte Carlo Tree
Search (MCTS) on a given code to improve performance. Each code 
has a corresponding history of transformations and predicted cost.
You are given the code of the selected node and its immediate
parent. Their main differences include:
\\

Loop shapes:

Current: \\
for b\_1, t\_1, j\_1 in T.grid(1, 2, 8), and then 

j\_2\_init in T.grid(1), and later 

j\_2 in T.grid(1).

Index example: vj = ... + j\_1 * 64 + j\_2 * 64

Parent: 

for b\_1, t\_1, j\_1 in T.grid(1, 2, 2),

then j\_2\_init in T.grid(4), 

and later j\_2 in T.grid(4).

Index example: vj = ... + j\_1 * 256 + j\_2 * 64\\

Tile decisions:

Current: sample\_perfect\_tile(..., decision=[4, 8, 1, 64])

Parent: sample\_perfect\_tile(..., decision=[4, 2, 4, 64])\\

Performance estimates:

Current: 0.773

Parent: 0.313\\

Available transformations:\\
TileSize, Parallel, ComputeLocation, Unroll\\

Task\\
Analyze the IR, trace, and predicted scores. \\
Then propose a sequence of transformations (you may repeat any) to 
potentially improve performance. \\
Output your reasoning and your suggested transformations.\\

For example, your answer should be in the following format:
Reasoning: This code still has large loop extents, so I'd tile it 
twice differently, then unroll.\\
Transformations to apply: TileSize, TileSize, Unroll.
\end{promptbox}

\begin{promptbox}[LLM Response]
Reasoning: The current schedule tiles the j-axis as 2048=4×8×1×64,
whereas the parent uses 4×2×4×64. The higher predicted score of 
the current node (0.773 vs. 0.313) suggests that expanding outer 
parallel granularity (j\_1=8) helped, but collapsing j\_2 to 1 
removed a small register-level tile, limiting unrolling and 
hurting reuse within each j\_3=64 chunk. \\

I would (1) reintroduce a non-unit j\_2 while shrinking j\_1 so that
j\_1×j\_2~=8; for instance, [4,4,2,64] keeps enough outer tasks but
creates a j\_2×j\_3=128-wide microtile that unrolls cleanly. 
(2) Tile the 7168-step k-reduction to align with j\_3=64, e.g., 
7168 = 7×16×64; the 16-step inner slice can be unrolled to cut 
loop overhead and improve register reuse across the j\_3 tile. 
(3) Move the compute of block "moe" to the j\_1 level so each j\_1
tile initializes/accumulates C locally, shortening live ranges and
reducing write-back pressure. 
(4) Parallelize the outer j\_1 (or t\_1 if j\_1 becomes too small) to 
utilize cores without touching the reduction. 
(5) Unroll the small j\_2 loop and the k-inner microtile. These 
changes retain the beneficial outer structure of the current node 
while restoring a compact inner tile for ILP and cache locality, 
which should further improve throughput.\\

Transformations to apply: TileSize, TileSize, ComputeLocation,
Parallel, Unroll, Unroll.
\end{promptbox}

\newpage

\section{Sample Efficiency and Speedup Results}
Table~\ref{tab:speedup-results} presents the relative speedup of three methods---\baseline, \mcts, and the \compiler---evaluated across the different benchmarks. 
Speedup is measured as the ratio of execution time for the unoptimized code to that of the optimized code after applying a given number of transformation proposals.
The table captures performance as a function of the number of samples explored. 
Higher values indicate more effective optimization.
For instance, the \compiler consistently achieves higher speedups with fewer samples, demonstrating superior sample efficiency and faster convergence compared to \mcts and \baseline.
This table corresponds to \fref{fig:improvement_a} in the paper.

\begin{table*}[ht]
\centering
\caption{Speedup over unoptimized code across varying numbers of samples for different compiler optimization methods.}
\label{tab:speedup-results}
\resizebox{\textwidth}{!}{%
\begin{tabular}{@{}llccccccccc@{}}
\toprule

\multirow{6}{*}{\llamathreebtable Attention Layer}
 & Number of Samples & 18 & 36 & 72 & 144 & 192 & 600 & 900 & 1632 & 5952 \\
\cmidrule{2-11}
 & \baselinetable     & 4.67 & 5.70 & 7.74 & 7.98 & 9.40 & 9.54 & 11.20 & 12.04 & 13.18 \\
 & \mctstable              & 4.14 & 4.68 & 8.11 & 8.50 & 9.66 & 9.94 & 11.79 & 12.44 & 12.63 \\
 & \compilertable        & 4.52 & 7.08 & 8.38 & 8.79 & 10.56 & 11.33 & 12.10 & 12.57 & 12.87 \\
\midrule

\multirow{6}{*}{\deepseekronetable MoE Layer}
 & Number of Samples & 36 & 54 & 72 & 144 & 192 & 600 & 900 & 1632 & 3000 \\
\cmidrule{2-11}
 & \baselinetable     & 2.11 & 2.27 & 3.90 & 4.10 & 5.07 & 6.60 & 6.62 & 9.31 & 9.13 \\
 & \mctstable              & 5.93 & 6.33 & 6.79 & 6.87 & 6.93 & 7.24 & 8.18 & 8.84 & 8.90 \\
 & \compilertable        & 7.05 & 7.33 & 8.34 & 8.53 & 9.10 & 9.45 & 11.06 & 11.74 & 11.74 \\
\midrule

\multirow{6}{*}{\fluxtable Attention Layer}
 & Number of Samples & 36 & 54 & 72 & 150 & 200 & 600 & 1000 & 1500 & 3000 \\
\cmidrule{2-11}
 & \baselinetable     & 2.22 & 2.44 & 2.73 & 2.73 & 4.64 & 4.71 & 5.11 & 5.61 & 5.58 \\
 & \mctstable              & 3.62 & 3.79 & 3.85 & 4.04 & 4.37 & 5.09 & 5.56 & 5.40 & 5.64 \\
 & \compilertable        & 4.48 & 4.67 & 4.89 & 5.37 & 5.42 & 5.43 & 5.59 & 5.60 & 5.67 \\
\midrule

\multirow{6}{*}{\fluxtable Convolution Layer}
 & Number of Samples & 36 & 72 & 150 & 200 &400 & 600 & 1000 & 1600 & 3000 \\
\cmidrule{2-11}
 & \baselinetable     & 2.08 & 2.11 & 2.15 & 2.19 & 2.29 & 2.32 & 2.44 & 2.45 & 2.55 \\
 & \mctstable              & 2.11  & 2.13 & 2.18 & 2.37 & 2.38 & 2.38 & 2.44 & 2.44 & 2.45 \\
 & \compilertable        & 2.21  & 2.30 & 2.29 & 2.36 & 2.47 & 2.47 & 2.51 & 2.55 & 2.58 \\
\midrule

\multirow{6}{*}{\llamafourscouttable MLP Layer}
 & Number of Samples & 20 & 50 & 100 & 250 &400 & 600 & 1000 & 1500 & 3000 \\
\cmidrule{2-11}
 & \baselinetable     & 1.36 & 2.28 & 3.61 & 5.59 & 5.59 & 5.75 & 5.76 & 5.94  & 5.94  \\
 & \mctstable              & 1.76  & 2.51 & 4.05 & 5.41 & 7.83 & 8.13 & 8.58 & 8.90 & 8.90 \\
 & \compilertable        & 12.74  & 12.74 & 12.74 & 12.75 & 12.75 & 13.24 & 13.26 & 13.52 & 13.79 \\
\bottomrule
\end{tabular}%
}
\end{table*}

\newpage

\section{Impact of LLM Choice and Reasoning Strategy}

As a continuation of \frefsub{fig:llm-ablation}{a}, \tref{tab:model-ablation} reports speedup over unoptimized code on three additional benchmarks: \deepseekbench, \fluxattbench, and \fluxconvbench. 
Each block of the table corresponds to a different benchmark and shows the best speedup achieved by the \compiler as a function of the number of schedules sampled using the reasoning model listed in the table. 
Rows compare different reasoning models used for API call generation, including both proprietary (e.g., GPT-4o mini, OpenAI o1-mini) and open-source models (e.g., Llama3.3-Instruct, DeepSeek-Distill).
Across all benchmarks, the results show that more capable models—those that are larger or instruction-tuned—consistently achieve higher speedups with fewer samples. 
For example, Llama3.3-Instruct (70B) and DeepSeek-Distill (32B) achieve near-maximal speedup within the first 72–150 samples, while smaller models such as DeepSeek-Distill (7B) or Llama3.1-Instruct (8B) reach similar performance more gradually. 
These results validate the generality of our findings: the use of context-aware LLMs accelerates convergence of the \compiler across diverse code domains. 
Moreover, the performance of open-source models is competitive with proprietary alternatives, further supporting the accessibility and reproducibility of our method.

\begin{table*}[ht]
\centering
\caption{Speedup over unoptimized code across varying numbers of samples for different choices of API call models.}
\label{tab:model-ablation}
\resizebox{\textwidth}{!}{%
\begin{tabular}{@{}llccccccccc@{}}
\toprule

\multirow{6}{*}{Llama3-8B Attention Layer}
 & Number of Samples & 18 & 36 & 72 & 150 & 200 & 600 \\
\cmidrule{2-8}
 & GPT-4o mini &4.52	&7.08	&8.38	&8.79	&10.56	&11.33\\
 & OpenAI o1-mini &4.63	&4.64	&7.37	&9.14	&9.15	&11.77 \\
 & Llama3.3-Instruct (70B)     &5.15	&9.68 &9.69 &9.80 &9.80	&9.81 \\
 & DeepSeek-Distill-Qwen (32B)    &7.07	 &8.14	&8.23	&8.77  &8.78 &9.98 \\
 & Llama3.1-Instruct (8B)     &3.60	&5.87	&6.28	&8.46	&8.63	&10.52 \\
 & DeepSeek-Distill-Qwen (7B) &4.06	&4.86	&6.68	&6.82	&7.94	&11.58 \\
\midrule

\multirow{6}{*}{\deepseekronetable MoE Layer}
 & Number of Samples & 18 & 36 & 72 & 150 & 200 & 600 \\
\cmidrule{2-8}
 & GPT-4o mini &6.14	&7.05	&8.33	&8.53	&9.10	&9.45\\
 & OpenAI o1-mini &4.56	&6.65	&8.59	&9.29	&10.55	&11.56 \\
 & Llama3.3-Instruct (70B) &7.30	&7.70	&7.96	&8.06	&8.60	&9.22    \\
 & DeepSeek-Distill-Qwen (32B)   &5.56	&8.11	&9.49	&10.17	&11.02	&12.02   \\
 & Llama3.1-Instruct (8B)    &4.29	&4.31	&6.98	&8.70	&9.18	&9.21    \\
 & DeepSeek-Distill-Qwen (7B)  &6.89	&7.35	&7.35	&10.22	&10.34	&10.44\\
\midrule

\multirow{6}{*}{\fluxtable Attention Layer}
 & Number of Samples & 18 & 36 & 72 & 150 & 200 & 600 \\
\cmidrule{2-8}
 & GPT-4o mini &4.09	&4.48	&4.89	&5.37	&5.42	&5.43\\
 & OpenAI o1-mini  &3.29	&2.99	&5.27	&5.53	&5.65	&5.67\\
 & Llama3.3-Instruct (70B)   &2.67	&3.12  &4.82	&4.86	&5.71	&5.71   \\
 & DeepSeek-Distill-Qwen (32B) &3.56	&4.29	&4.29	&4.54	&4.99 &5.21           \\
 & Llama3.1-Instruct (8B)     &2.01	&3.43	&3.55	&3.80	&3.87 &5.21    \\
 & DeepSeek-Distill-Qwen (7B)  &3.02	&3.76	&3.83	&4.54	&4.94 &5.17\\
\midrule

\multirow{6}{*}{\fluxtable Convolution Layer}
 & Number of Samples & 18 & 36 & 72 & 150 & 200 & 600 \\
\cmidrule{2-8}
 & GPT-4o mini &1.65	&2.21	&2.30	&2.29	&2.36	&2.47\\
 & OpenAI o1-mini  &2.37	&2.37	&2.38	&2.39	&2.45	&2.54\\
 & Llama3.3-Instruct (70B)   &2.30	&2.35	&2.47	&2.51	&2.56	&2.57  \\
 & DeepSeek-Distill-Qwen (32B)   &1.41	&2.26	&2.32	&2.35	&2.40	&2.45         \\
 & Llama3.1-Instruct (8B) &2.11	&2.30	&2.39	&2.55	&2.55	&2.56        \\
 & DeepSeek-Distill-Qwen (7B)  &1.56	&2.18	&2.42	&2.44	&2.46	&2.45\\
\bottomrule
\end{tabular}%
}
\end{table*}

\newpage

\section{Impact of Historical Trace Depth on Optimization Efficiency}
As a continuation of \frefsub{fig:llm-ablation}{b}, \tref{tab:context-ablation} presents the data for the ablation study on the depth of historical trace included in the prompt sent to the LLM.
Specifically, we compare two configurations: the “Parent + Grandparent” setting, where the prompt contains information from the current node and its two immediate ancestors, and the “Parent + Grandparent + Great-Grandparent” setting, where the prompt additionally includes the great-grandparent node. 
These variations allow us to assess the impact of deeper context windows on the effectiveness of the \compiler.

Results show that increasing the historical context generally improves sample efficiency across all benchmarks. 
For example, on \deepseekbench, adding one more ancestral node boosts early performance significantly, achieving a 9.39$\times$ speedup at just 18 samples compared to 6.14$\times$ for the shallower context. 
Similarly, on \llamabench, the extended context leads to a higher final speedup (11.87$\times$ vs. 11.33$\times$) and earlier convergence. 
The performance gains, while smaller, are also consistent on \fluxattbench and \fluxconvbench, with improvements observed across all sample budgets.
These findings confirm that providing richer historical context enables the LLM to make more informed decisions at each step of the search, ultimately enhancing the sample efficiency of the \compiler

\begin{table*}[ht]
\centering
\caption{Speedup over unoptimized code across varying numbers of samples for different context lengths.}
\label{tab:context-ablation}
\resizebox{\textwidth}{!}{%
\begin{tabular}{@{}llccccccccc@{}}
\toprule

\multirow{3}{*}{\llamathreebtable Attention Layer}
 & Number of Samples & 18 & 36 & 72 & 150 & 200 & 600 \\
\cmidrule{2-8}
 & Parent + Grandparent    &4.52	&7.08	&8.38	&8.79	&10.56	&11.33  \\
 & Parent + Grandparent + Great-Grandparent    &3.63	&7.13	&11.36	&11.86	&11.86	&11.87        \\
\midrule

\multirow{3}{*}{\deepseekronetable MoE Layer}
 & Number of Samples & 18 & 36 & 72 & 150 & 200 & 600 \\
\cmidrule{2-8}
 & Parent + Grandparent     &6.14	&7.05	&8.33	&8.53	&9.10	&9.45 \\
 & Parent + Grandparent + Great-Grandparent     &9.39	&10.31	&10.31	&10.49	&10.59	&10.65     \\
\midrule

\multirow{3}{*}{\fluxtable Attention Layer}
 & Number of Samples & 18 & 36 & 72 & 150 & 200 & 600 \\
\cmidrule{2-8}
 & Parent + Grandparent     &4.09	&4.48	&4.89	&5.37	&5.42	&5.43\\
 & Parent + Grandparent + Great-Grandparent     &4.21	&4.55	&4.81	&5.47	&5.53	&5.61      \\
\midrule

\multirow{3}{*}{\fluxtable Convolution Layer}
 & Number of Samples & 18 & 36 & 72 & 150 & 200 & 600 \\
\cmidrule{2-8}
 & Parent + Grandparent     &1.65	&2.21	&2.30	&2.29	&2.36	&2.47 \\
 & Parent + Grandparent + Great-Grandparent     &1.73	&2.22	&2.32	&2.35	&2.49	&2.50    \\
\bottomrule
\end{tabular}%
}
\end{table*}

\newpage

\section{Ablations of MCTS Branching Factor}
To determine the value of MCTS branching factor (\(B\)), we ablate on \(B=2\) and \(B=4\). In \tref{tab:branching}, results show that when branching factor \(B=2\), the \compiler is more sample-efficient than when \(B=4\). Our choice of \(B=2\) aligns with prior works ~\cite{branching1:cg:2007, ucb:ml:2002}. If a higher branching factor is chosen, then there are more possible next steps, which require more sampling effort (i.e., more simulations) to cover these expanded possibilities at the same level of thoroughness.
\begin{table*}[ht]
\centering
\caption{Speedup over unoptimized code across varying numbers of samples for different branching factors.}
\label{tab:branching}
\resizebox{\textwidth}{!}{%
\begin{tabular}{@{}llccccccccc@{}}
\toprule

\multirow{3}{*}{\llamathreebtable Attention Layer}
 & Number of Samples & 18 & 36 & 72 & 150 & 200 & 600 \\
\cmidrule{2-8}
 & B = 2    &4.52	&7.08	&8.38	&8.79	&10.56	&11.33  \\
 & B = 4     &4.16	&7.88	&8.35	&8.89	&9.86	&10.99       \\
\midrule

\multirow{3}{*}{\deepseekronetable MoE Layer}
 & Number of Samples & 18 & 36 & 72 & 150 & 200 & 600 \\
\cmidrule{2-8}
 & B = 2     &6.14	&7.05	&8.33	&8.53	&9.10	&9.45 \\
 & B = 4     &2.98	&4.29	&4.29	&7.28   &7.29	&9.10     \\
\midrule

\multirow{3}{*}{\fluxtable Attention Layer}
 & Number of Samples & 18 & 36 & 72 & 150 & 200 & 600 \\
\cmidrule{2-8}
 & B = 2     &4.09	&4.48	&4.89	&5.37	&5.42	&5.43\\
 & B = 4    &2.40   &3.48	&3.97	&4.95	&4.97	&5.55       \\
\midrule

\multirow{3}{*}{\fluxtable Convolution Layer}
 & Number of Samples & 18 & 36 & 72 & 150 & 200 & 600 \\
\cmidrule{2-8}
 & B = 2     &1.65	&2.21	&2.30	&2.29	&2.36	&2.47 \\
 & B = 4     &1.91	&1.97	&2.23	&2.23	&2.25	&2.43   \\
\bottomrule
\end{tabular}%
}
\end{table*}

\newpage
\section{Cost of LLMs Used in Experiments}
In ~\tref{tab:cost-per-experiment}, for each benchmark, we report the API cost of running a full experiment with every LLM used to generate transformation proposals. 
We run a high number of samples to understand the boundary of performance improvements and allow the algorithm to saturate. 
For OpenAI, our main results used GPT-4o mini, the lowest-cost model available at submission time. For open-source models, we used Hugging Face APIs through the Nscale hyperscaler provider. 
Across benchmarks, these open-source models achieved competitive speedups and sample efficiency relative to GPT-4o mini, indicating that open-source models are a viable alternative when commercial APIs are impractical. 
Costs of open-source models could be further reduced by local deployment.
\begin{table}[ht]
  \centering
  \small 
  \setcellgapes{-2.5pt}
  \makegapedcells
  \setlength{\tabcolsep}{3.4pt} 
  \captionsetup{skip=7pt}
  \caption{Cost of different LLM APIs per entire experiment (USD) across layer-wise and end-to-end benchmarks.}
  \label{tab:cost-per-experiment}
  \begin{tabular}{@{}lrrrrrr@{}}
    \toprule
    \multicolumn{1}{c}{\multirow{2}{*}[-2ex]{\textbf{\makecell{Layer / Task}}}} &
    \multicolumn{6}{c}{\textbf{Model}} \\
    \cmidrule(lr){2-7}
    &
    \textbf{\makecell{GPT-4o \\ mini}} &
    \textbf{\makecell{OpenAI \\ o1-mini}} &
    \textbf{\makecell{Llama3.3-\\Instruct (70B)}} &
    \textbf{\makecell{DeepSeek-\\Distill (32B)}} &
    \textbf{\makecell{Llama3.1-\\Instruct (8B)}} &
    \textbf{\makecell{DeepSeek-\\Distill (7B)}} \\
    \midrule
    \llamathreebtable Attention Layer                    & \$0.89 & \$6.56 & \$2.07 & \$1.55 & \$0.31 & \$2.07 \\
    \deepseekronetable MoE Layer                         & \$0.90 & \$6.63 & \$2.09 & \$1.57 & \$0.31 & \$2.09 \\
    \fluxtable Attention Layer                          & \$0.88 & \$6.47 & \$2.03 & \$1.52 & \$0.30 & \$2.03 \\
    \fluxtable Convolution Layer                             & \$1.12 & \$8.25 & \$2.67 & \$2.00 & \$0.40 & \$2.67 \\
    Llama-4-Scout MLP Layer & \$0.90 & ---   & ---   & ---   & ---   & ---   \\
    \makecell[l]{End-to-End \llamathreebtable}   & \$1.59 & ---   & ---   & ---   & ---   & ---   \\
    \bottomrule
  \end{tabular}
\end{table}

\section{LLM Proposal Validity and Fallback Rates}

LLM‑generated transformations can occasionally be syntactically valid but semantically redundant or performance-regressive. 
During any single MCTS expansion, proposals that fail basic validity checks (e.g., naming or use‑context non‑compliance) are simply discarded while the remaining valid proposals proceed, and no fallback is triggered. 
A fallback occurs only when all LLM-generated proposals in that expansion are invalid, in which case the search reverts to the default, non‑LLM expansion policy and continues without interruption. 
In \tref{tab:fallback-rates}, we report the fallback rate as the average fraction of expansions that trigger this non‑LLM path (i.e., expansions in which all LLM proposals are invalid). 
To prevent downstream harm from poor but valid transformations, the cost model evaluates all proposed transformations before they are added to the tree; proposals with low estimated values are naturally pruned.
Because the transformation space is a known, finite set of legal rewrites, most correctness issues reduce to naming compliance and use‑context, which modern instruction‑tuned LLMs typically handle well.
Empirically, commercial models (GPT‑4o mini and OpenAI o1‑mini) show 0\% fallback rates, larger open‑source models perform similarly (Llama3.3‑Instruct 70B at 0.08\% and DeepSeek‑Distill 32B at 0.17\%), whereas smaller models exhibit higher fallback rates (Llama3.1‑Instruct 8B at 10.50\% and DeepSeek‑Distill 7B at 17.20\%).

\begin{table}[ht]
  \centering
  \small 
  \setcellgapes{-2.5pt}
  \setlength{\tabcolsep}{8.5pt} 
  \captionsetup{skip=7pt}
  \caption{Fallback rate by model used as the transformation proposal generator.}
  \label{tab:fallback-rates}

  \begin{tabular}{@{}lr@{}}
    \toprule
    \textbf{Model} & \textbf{Fallback Rate} \\
    \midrule
    GPT-4o mini & 0\% \\
    OpenAI o1-mini & 0\% \\
    Llama3.3-Instruct (70B) & 0.08\% \\
    DeepSeek-Distill (32B) & 0.17\% \\
    Llama3.1-Instruct (8B) & 10.50\% \\
    DeepSeek-Distill (7B) & 17.20\% \\
    \bottomrule
  \end{tabular}
\end{table}

\newpage
\section*{NeurIPS Paper Checklist}

\begin{enumerate}

\item {\bf Claims}
    \item[] Question: Do the main claims made in the abstract and introduction accurately reflect the paper's contributions and scope?
    \item[] Answer: \answerYes{} 
    \item[] Justification: We demonstrate accurately the paper's contributions and scope in the abstract, \sref{sec:introduction} (introduction), and \sref{sec:results} (results) to support the claims.

\item {\bf Limitations}
    \item[] Question: Does the paper discuss the limitations of the work performed by the authors?
    \item[] Answer: \answerYes{} 
    \item[] Justification: The authors acknowledge that their method currently depends on external APIs for querying large language models, which may pose reproducibility and scalability concerns due to cost and access restrictions. They also recognize that the system’s performance can vary across model types and that the evaluation is limited to six representative state-of-the-art benchmarks. Moreover, since the approach relies on prompt formatting and reasoning traces, its effectiveness may degrade in settings where context length or LLM interpretability is constrained.

\item {\bf Theory assumptions and proofs}
    \item[] Question: For each theoretical result, does the paper provide the full set of assumptions and a complete (and correct) proof?
    \item[] Answer:  \answerNA{}{} 
    \item[] Justification: This paper focuses on practical compiler optimization techniques rather than theoretical developments. As such, it does not present formal theorems or proofs. However, in \sref{sec:formalization}, we provide a formal problem formulation to clearly define the optimization setting and guide our methodology. No theoretical claims are made that would require formal assumptions or correctness proofs.

    \item {\bf Experimental result reproducibility}
    \item[] Question: Does the paper fully disclose all the information needed to reproduce the main experimental results of the paper to the extent that it affects the main claims and/or conclusions of the paper (regardless of whether the code and data are provided or not)?
    \item[] Answer: \answerYes{} 
    \item[] Justification: We have provided a detailed experimental setup and included the link to our GitHub repository. We also described in detail our method in \sref{sec:llm-reasoning} and \sref{sec:mcts} to make sure our experiments can be reproduced.

\item {\bf Open access to data and code}
    \item[] Question: Does the paper provide open access to the data and code, with sufficient instructions to faithfully reproduce the main experimental results, as described in supplemental material?
    \item[] Answer: \answerYes{} 
    \item[] Justification: We included the link to our \href{https://github.com/Anna-Bele/LLM_MCTS_Search}{repository} in the abstract. The repository contains instructions on how to set up and run the experiments.

\item {\bf Experimental setting/details}
    \item[] Question: Does the paper specify all the training and test details (e.g., data splits, hyperparameters, how they were chosen, type of optimizer, etc.) necessary to understand the results?
    \item[] Answer: \answerYes{} 
    \item[] Justification: In \sref{sec:experimental_setup}, we specified all the experiment details necessary to understand the results.

\item {\bf Experiment statistical significance}
    \item[] Question: Does the paper report error bars suitably and correctly defined or other appropriate information about the statistical significance of the experiments?
    \item[] Answer: \answerYes{} 
    \item[] Justification: All experiments are repeated 20 times, and the results are averaged to ensure statistical stability, as described in \sref{sec:experimental_setup}.

\item {\bf Experiments compute resources}
    \item[] Question: For each experiment, does the paper provide sufficient information on the computer resources (type of compute workers, memory, time of execution) needed to reproduce the experiments?
    \item[] Answer: \answerYes{}
    \item[] Justification: We mention the machine details in \sref{sec:experimental_setup}, and the README in the GitHub repository provides the steps.
    
\item {\bf Code of ethics}
    \item[] Question: Does the research conducted in the paper conform, in every respect, with the NeurIPS Code of Ethics \url{https://neurips.cc/public/EthicsGuidelines}?
    \item[] Answer: \answerYes{}
    \item[] Justification: The research conducted in the paper conforms, in every respect, with the NeurIPS Code of Ethics.

\item {\bf Broader impacts}
    \item[] Question: Does the paper discuss both potential positive societal impacts and negative societal impacts of the work performed?
    \item[] Answer: \answerYes{} 
    \item[] Justification: This work presents a compiler optimization framework that leverages LLM reasoning for efficient model serving. The positive societal impacts include reducing the computational cost of deploying large machine learning models, which in turn improves accessibility and scalability, as discussed in abstract, introduction, and conclusion (see \sref{sec:introduction} and \sref{sec:conclusion}).
    
\item {\bf Safeguards}
    \item[] Question: Does the paper describe safeguards that have been put in place for responsible release of data or models that have a high risk for misuse (e.g., pretrained language models, image generators, or scraped datasets)?
    \item[] Answer:  \answerNA{}
    \item[] Justification: The paper does not release any models or associated datasets which have high risk of misuse. It rather focuses on compiler-level optimizations for efficient ML model serving, which poses no direct safety or misuse concerns that would warrant safeguards.

\item {\bf Licenses for existing assets}
    \item[] Question: Are the creators or original owners of assets (e.g., code, data, models), used in the paper, properly credited and are the license and terms of use explicitly mentioned and properly respected?
    \item[] Answer: \answerYes{} 
    \item[] Justification: Our method is integrated with Apache TVM v0.20.0~\cite{tvm}, an open-source machine learning compiler stack released under the Apache License 2.0. We properly cite the original work~\cite{tvm, metaschedule:neurips:2022} and ensure full compliance with its licensing terms. We also use OpenAI or HuggingFace's model serving and utilize their APIs to access the models.

\item {\bf New assets}
    \item[] Question: Are new assets introduced in the paper well documented and is the documentation provided alongside the assets?
    \item[] Answer: \answerYes{} 
    \item[] Justification: We integrate the proposed approach into open-source TVM scheduling and also make our code open source, as discussed in \sref{sec:experimental_setup}.

\item {\bf Crowdsourcing and research with human subjects}
    \item[] Question: For crowdsourcing experiments and research with human subjects, does the paper include the full text of instructions given to participants and screenshots, if applicable, as well as details about compensation (if any)? 
    \item[] Answer: \answerNA{}
    \item[] Justification: The paper does not involve crowdsourcing nor research with human subjects.

\item {\bf Institutional review board (IRB) approvals or equivalent for research with human subjects}
    \item[] Question: Does the paper describe potential risks incurred by study participants, whether such risks were disclosed to the subjects, and whether Institutional Review Board (IRB) approvals (or an equivalent approval/review based on the requirements of your country or institution) were obtained?
    \item[] Answer: \answerNA{} 
    \item[] Justification: The paper does not involve crowdsourcing nor research with human subjects.

\item {\bf Declaration of LLM usage}
    \item[] Question: Does the paper describe the usage of LLMs if it is an important, original, or non-standard component of the core methods in this research? Note that if the LLM is used only for writing, editing, or formatting purposes and does not impact the core methodology, scientific rigorousness, or originality of the research, declaration is not required.
    \item[] Answer: \answerYes{} 
    \item[] Justification: Large language models are an integral part of our method. We use them in the program optimization process to guide transformation proposals in compiler optimization search. This use of LLMs is central, and is described in detail in \ref{sec:llm-reasoning}. All of our usage complies with responsible AI guidelines, and models used (e.g., OpenAI's models, LLaMA-3, DeepSeek) are publicly accessible using APIs.

\end{enumerate}

\end{document}